\documentclass[preprint,12pt]{elsarticle}
\usepackage{amsmath}
\usepackage{algorithm, algorithmicx, algpseudocode}

\usepackage{multirow}
\usepackage{makecell}
\usepackage{hyperref}
\usepackage{graphicx}
\usepackage{amssymb}
\journal{Elsevier}

\begin{document}

\begin{frontmatter}

%% Title, authors and addresses

%% use the tnoteref command within \title for footnotes;
%% use the tnotetext command for theassociated footnote;
%% use the fnref command within \author or \address for footnotes;
%% use the fntext command for theassociated footnote;
%% use the corref command within \author for corresponding author footnotes;
%% use the cortext command for theassociated footnote;
%% use the ead command for the email address,
%% and the form \ead[url] for the home page:
%% \title{Title\tnoteref{label1}}
%% \tnotetext[label1]{}
%% \author{Name\corref{cor1}\fnref{label2}}
%% \ead{email address}
%% \ead[url]{home page}
%% \fntext[label2]{}
%% \cortext[cor1]{}
%% \affiliation{organization={},
%%             addressline={},
%%             city={},
%%             postcode={},
%%             state={},
%%             country={}}
%% \fntext[label3]{}

\title{Multi-stages attention breast cancer classification based on nonlinear spiking neural P neurons with autapses}
\tnotetext[mytitlenote]{}

%% use optional labels to link authors explicitly to addresses:
%% \author[label1,label2]{}
%% \affiliation[label1]{organization={},
%%             addressline={},
%%             city={},
%%             postcode={},
%%             state={},
%%             country={}}
%%
%% \affiliation[label2]{organization={},
%%             addressline={},
%%             city={},
%%             postcode={},
%%             state={},
%%             country={}}

\author[inst1]{Bo Yang}
\author[inst1]{Hong Peng\corref{cauthor}}
\author[inst1]{Xiaohui Luo}
\author[inst2]{Jun Wang}
\cortext[cauthor]{Corresponding author}
\affiliation[inst1]{organization={School of Computer and Software Engineering},%Department and Organization
            addressline={Xihua University}, 
            city={Chengdu},
            postcode={610039}, 
            state={Sichuan},
            country={China}}
\affiliation[inst2]{organization={School of Electrical Engineering and Electronic Information},%Department and Organization
            addressline={Xihua University}, 
            city={Chengdu},
            postcode={610039}, 
            state={Sichuan},
            country={China}}
\begin{abstract}
%% Text of abstract
Breast cancer(BC) is a prevalent type of malignant tumor in women. 
Early diagnosis and treatment are vital for enhancing the patients' survival rate. Downsampling in deep networks may lead to loss of information, so for compensating the detail and edge information and allowing convolutional neural networks to pay more attention to seek the lesion region, we propose a multi-stages attention architecture based on NSNP neurons with autapses. First, unlike the single-scale attention acquisition methods of existing methods, we set up spatial attention acquisition at each feature map scale of the convolutional network to obtain an fusion global information on attention guidance. Then we introduce a new type of NSNP variants called NSNP neurons with autapses. Specifically, NSNP systems are modularized as feature encoders, recoding the features extracted from convolutional neural network as well as the fusion of attention information and preserve the key characteristic elements in feature maps. This ensures the retention of valuable data while gradually transforming high-dimensional complicated info into low-dimensional ones. The proposed method is evaluated on the public dataset BreakHis at various magnifications and classification tasks. It achieves a classification accuracy of 96.32\% at all magnification cases, outperforming state-of-the-art methods. Ablation studies are also performed, verifying the proposed model's efficacy. The source code is available at \href{https://github.com/XhuBobYoung/Breast-cancer-Classification.git}{XhuBobYoung/Breast-cancer-Classification}.
\end{abstract}

%%Graphical abstract

%%Research highlights


\end{frontmatter}

%% \linenumbers

%% main text
\section{Introduction}\label{sec1}

`Breast cancer is a widespread malignancy among women worldwide, which demands timely diagnosis and treatment for improved survival and cure rates. Computer-assisted therapy for breast cancer is a revolutionary method that employs advanced computer technology to aid doctors in treating the disease. 
Through analyzing mammograms, ultrasound, and MRI images, this technology assists doctors in accurately diagnosing the location, size, morphology, and other characteristics of the lesion. It provides a strong foundation for treatment plan design and evaluation of treatment effectiveness.
In the past few years, traditional approaches to breast cancer classification relied on manually designed feature extraction and classifier. Classification techniques include support vector machines and artificial neural networks. 
These methods are often subject to inadequate feature extraction and subjective classifier design, leading to suboptimal classification results. 
However, with the advent of deep learning techniques, it has become possible to extract features from images automatically and classify them into malignant and benign categories. 
Despite the promising results of deep learning-based methods, certain issues such as over-fitting and unbalanced datasets persist. 
\begin{figure}
    \centering
    \includegraphics[width=3in]{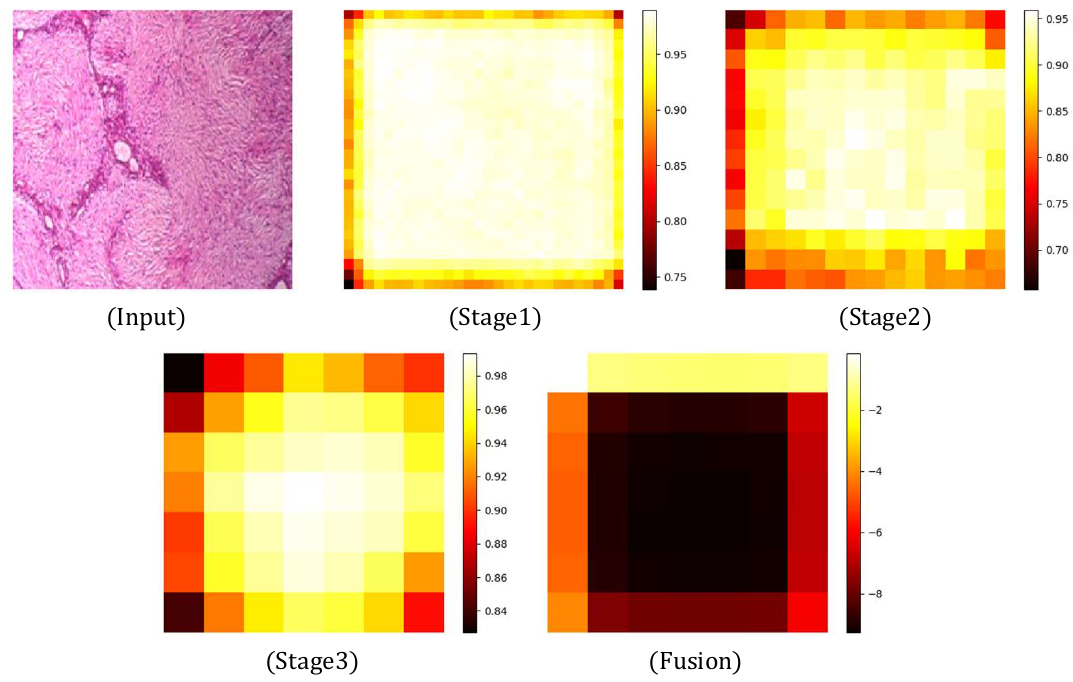}
    \caption{Attention heat map at different stages. As the network deepens, the downsampling operation makes the image lower in resolution, while losing quite a lot of detail information. Stage 2 and stage 3 show less detail than stage 1. when all the attention maps are finally fused, the attention image shows a completely different state.}
    \label{fig1}
\end{figure}
\subsection{Related work}

\subsubsection{Breast cancer classification}
In the field of deep learning, transfer learning has become an effective technique to improve the performance of machine learning models. 
The primary implementation approach of transfer learning involves two steps: optimizing large networks pre-trained on massive datasets, and fine-tuning on dedicated datasets. 
It allows knowledge and patterns learned from large datasets to be migrated to small datasets. This results in improved model accuracy and efficiency.
Transfer learning has been particularly relevant in situations where the availability of large datasets is limited. The cost of training a new model from scratch is high. 
As a result, pre-trained models have become popular due to their superior performance and versatility at medical images processing. 
For instance, Vesal et al.~\cite{vesal2018classification} normalized a portion of the images from the BACH 2018 grand challenge by taking extracted image patches to feed the images into Inception-V3 and ResNet50, which have been trained on ImageNet. 
Deepak et al.~\cite{deepak2019brain} used pre-trained GoogLeNet to extract brain MRI image features, which achieves patient-level five-fold cross-validation, and an average accuracy of 98\% is achieved in the triple classification of brain tumors. 
To test the effect of different normalization methods on the recognition rate of the model, Kassani et al.~\cite{kassani2019breast} conducted experiments on the BACH dataset using the pre-trained model Xception. 
As a result, an average accuracy of 92.50\% was achieved. 
Khan et al.~\cite{khan2019novel} classified breast cancer programs with 3 different pre-trained models, GoogLeNet, VGG, and ResNet, whose average recognition accuracy was 97.53\%. 
Most of the pre-trained models are based on natural datasets and are not specific to the medical field. 
Based on this, Alzubaidi et al.~\cite{alzubaidi2020optimizing} first pre-trained with several same area datasets, and then trained the target dataset based on this to achieve an unusually classification effect. 
Hassan et al.~\cite{hassan2020breast} compared two pre-trained models, AlexNet and GoogLeNet, on CBIS-DDSM and INbreast datasets proving that AlexNet performs better and has stronger robustness. 
Celik et al.~\cite{celik2020automated} achieved a more promising recognition accuracy on the BreakHis dataset by training the last few layers of ResNet50 and DenseNet161. 
Since automated breast ultrasound imaging has different views, Wang et al.~\cite{wang2020breast} proposed a CNN for multi-view feature extraction based on this property, which can extract more lesion information from various aspects. 
Based on a blockwise fine-tuning strategy, Boumaraf et al.~\cite{boumaraf2021new} designed a residual block and applied it to the pre-trained model ResNet18. In this way, the pre-training parameters of the ResNet used for initialization are only required to provide a process of computation and are not involved in back-propagation, thus effectively preventing overfitting. 
Karthiga et al.~\cite{karthiga2021transfer} combined with one-hot encoding technology and transfer learning, a 98.62\% accuracy rate was achieved on the BreakHis data set. 
Alruwaili et al.~\cite{alruwaili2022automated} aimed to improve the performance and efficiency of breast cancer detection on the Mias dataset by utilizing the Resnet-MobileNet architecture and implementing various training strategies. 
Segmented lesions were utilized by Saber et al. \cite{saber2021novel} as a preprocessing technique for mammograms, which were then combined with VGG-extracted features and further improved for the ultimate classification. 
Aljuaid et al.~\cite{aljuaid2022computer} aimed to enhance the accuracy of breast pathogram recognition by further improving the binary classification accuracy of ResNet, InceptionV3, and ShuffleNet~\cite{zhang2018shufflenet} on the BreakHis dataset, as well as enhancing its subdivision into multi-classification problems. However, while the above methods absorb the advantages of high accuracy and easy training of the pre-training model. Yet, they only continue to utilize the feature extraction results of the pre-trained model, while the middle feature information is ignored. This leads to a result that too much edge information and details are dropped during downsampling. As shown in Figure~\ref{fig1}, when images are fed into the network, a lot of detail information is reduced due to the reduced resolution, although the attention heat map at each stage has almost the same heat distribution. However, when adopting adaptive fusion, the attention heat map shows a different picture. The energy is inverted, while the transition between different regions is smoothed. Several methods take advantage of attention to guide the network to focus on importance, but they almost always make use of the attention information of stage 3. In fact, lesions are often tiny details and edge information, while stage 3 has filtered out a lot of the essential stuff.
\subsubsection{Attention mechanism}
Attention mechanisms have become a popular technique in deep learning. It enables a model to automatically select important information from a complicated sequence or image. The weights of channels and spatial can be learned through attention block , enhancing the model’s performance on focusing. 
In image-related tasks, the attention mechanism refers to the weighting of attention to different regions in an image to identify areas that are more relevant to the task, such as image classification and target detection. 
For instance, Wang et al.~\cite{wang2017residual} introduced the attention mechanism into the residual blocks for natural image classification. 
Hu et al.~\cite{hu2018squeeze} proposed Squeeze-and-Excitation (SE) for acquiring attention weight information between different channels. 
Woo et al.~\cite{woo2018cbam} integrated channel attention and spatial attention, combining the information from both for computer vision in classification and detection tasks. 
In the field of medical images, Schlemper et al.~\cite{schlemper2019attention} proposed U-Net with a gated attention mechanism, which enables the analysis of salient medical information in the feature stitching phase of the decoder. 
Kar et al.~\cite{karthik2022classification} integrated channel, spatial attention into two parallel CNNs to improve the segmentation capability of the model via an ensemble learning way.
\subsection{Motivation and contributions}
Pre-trained models, for its easy training properties, effectively avoids the problem of difficult convergence of models in the field of medical images due to too small a dataset. However, due to downsampling, there is a significant loss of edge information in images, and again details are difficult to recover. Therefore, to obtain the information in the middle layer of the pre-trained model, we extracted it employing spatial attention. Specifically, at each different scale of feature extraction stage, we extracted the feature attention information, and finally fused the information of different scales by convolution instead of using downsampling. In this way, the loss of information can be effectively avoided. Spiking neural P (SNP) systems~\cite{ionescu2006spiking} are neural-like computational models, which are inspired by the mechanisms of spiking neurons. 
Nonlinear spiking neural P (NSNP) systems~\cite{peng2020nonlinear} are nonlinear variants of SNP systems. NSNP systems have been successfully applied to various tasks due to the nonlinear spiking mechanisms, such as edge detection~\cite{rxian-lpylw22,jyan-zlpw22},  time series prediction~\cite{long-lpylws22,long-lpwy22,long-lxlpw22} and sentiment analysis~\cite{yhuang-plywop23,yhuang-lpwyo23,qliu-hypw23}. A promising result was achieved by yang et al. when they consider NSNP-type convolution on fundus images. However, NSNP-type convolution does not have a memory property, which may result in poor network performance when applied to deeper networks. To address this problem, we propose a variant of NSNP with memory, which is called NSNP neurons with autapses. NSNP neurons with autapses are modularized and embed into the pre-trained network. The process of recoding and classification is implemented at the same time on this module. The contribution of this paper can be summarized as follows:
\begin{enumerate}
\item[(1)] We propose multi-stages attention (MSA) frameworks, capturing attention information from various scale stages of convolutional neural networks. Both spatial and channel attention information are fused to guide the feature maps coding. This multi-scale approach can effectively compensate for the loss of edge and detail information due to downsampling.
\item[(2)] We introduce NSNP neurons with autapses. Inspired by this biological model, we construct an NSNP module, which is able to guarantee the preservation of key characteristic elements in feature maps, and progressively parses high-dimensional complex information into low-dimensional ones for classification.
\item[(3)] Based on the characteristics of the BreakHis dataset, we conducted systematic experiments to analyze and implement ablation. With different categorization tasks or resolutions, our method achieves more competitive results. Particularly in the 8 classification, we reached the highest metric among the state-of-the-art methods. And comprehensive analysis has be proved on the experiment results, demonstrating the effectiveness of multi-stages attention and NSNP modules.
\end{enumerate}

The remainder of this paper is organized as follows.
Section~\ref{sec2} introduces NSNP neurons with autapses.
Section~\ref{sec3} describes the proposed method for breast cancer classification in detail.
The experimental results are provided in Section~\ref{sec4}.
Finally, the conclusions are drawn in Section~\ref{sec5}.

%The remainder of this paper is organized as follows.
%Section~\ref{sec2} describes in detail the proposed detection method based on DTNP systems for microcalcification clusters in mammograms.
%The experiment and analysis are presented in Section~\ref{sec3}.
%Finally, the conclusions are drawn in Section~\ref{sec4}.

\section{NSNP neurons with autapses}\label{sec2}
\begin{figure}
    \centering
    \includegraphics[width=3in]{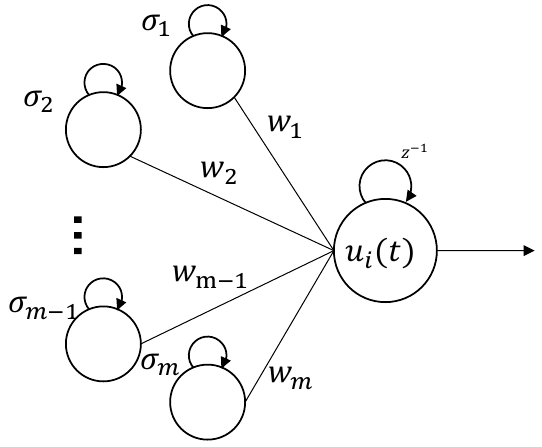}
    \caption{An NSNP neuron with autapses. $\sigma_1,\dots, \sigma_m$ represent m predecessor neurons, and $w_1,\dots, w_m$ represent the corresponding weight. $z^{-1}$ denotes an adaptive transformation of the state $u_i(t)$.}
    \label{fig2}
\end{figure}
An NSNP with autapses system has several neurons, which can be  described as
\begin{equation}
    (\sigma_1, \sigma_2, \dots , \sigma_m), m \in R
\end{equation}
Spiking neurons transmit information in the system in the form of firing in NSNP with autapses systems. $T$, $f(\cdot)$ and $g(\cdot)$ are fires threshold, generating nonlinear functions and consuming nonlinear functions, respectively.
Suppose the state value of neuron $\sigma_i, (1 \le i \le m)$ at moment $t$ is $u_i(t)$, when firing conditions $u_i(t)\ge T$ and $u_i(t)\ge g(u_i(t))$ both are satisfied, neuron $\sigma_i$ consumes the spikes of value $g(u_i(t))$ and generates the spikes of value $f(u_i(t))$. 
The generated spikes of value $f(u_i(t))$ will be sent into neuron $\sigma_j, (1 \le i \le m, j \ne i)$.

Based on the above firing rules, we can express the states of the neurons in the NSNP with autapses systems through an equation. 
For neuron $\sigma_i$ matching firing condition, its state $u_i(t)$ can be expressed as follows:
\begin{equation}
    u_i(t) = u_i(t-1)-g(u_i(t-1))+\sum^m_{i=1}w_i \times f(u_i(t-1))
    \label{eq3}
\end{equation}
where $g(\cdot)$ and $f(\cdot)$ represent the consumption and generation rules for neuron $\sigma_i$, respectively. 
It is worth mentioning that $g(\cdot)$ and $f(\cdot)$ can be represented by linear or nonlinear functions. 
$w_i$ denotes the connection weignt, and the spike of $u_i(t-1)$ received is the weighted sum of the $m$ predecessor neurons $\sigma_i$. 
As shown in Figure~\ref{fig2}, $u_i(t)$ represents the state of neuron $\sigma_i$ at time $t$. 
Each neuron has an autapses, denoted by $z^{-1}$. 
An NSNP neuron with autapses can receive and send spikes. 
Besides, when an NSNP neuron with autapses generates spikes, it consumes a certain amount of its own value of state. 
NSNP neurons with autapses in convolutional networks can be implemented straightforwardly. 
In convolutional networks, for example, we implement it via shifting the sequence of convolution and activation. 
The convolution equation inspired by NSNP systems can be summarized as follows:
\begin{equation}
    u_{out} = W_\lambda 
* \delta(\varrho(u_{in}))
\end{equation}
where $u_{in}$ and $u_{out}$ are the input feature maps and output feature maps, respectively. It begins with $u_{in}$ by batch normalization $\varrho$ and then activated by the activation function $\delta$. $*$ stands convolution process. $W_\lambda$ is a matrix of size $\lambda \times \lambda$, which is used to represent the convolutional weight.

\section{Method}\label{sec3}

\subsection{Proposed network architecture}

\begin{figure}
    \centering
    \includegraphics[width=4.8in]{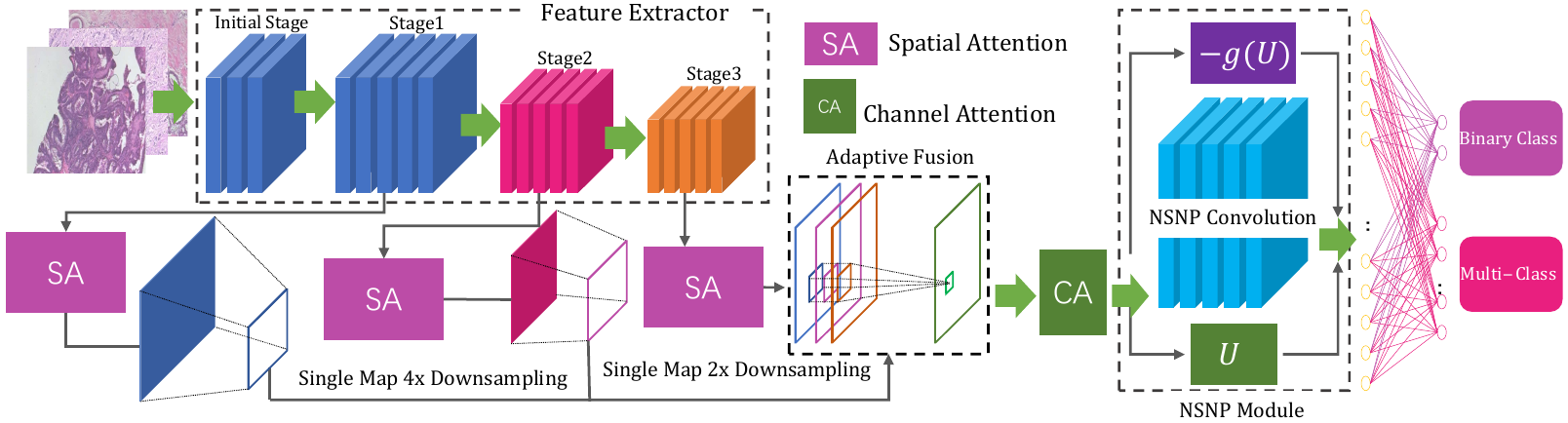}
    \caption{The architecture of the proposed network. SA and CA denote spatial and channel attention modules.}
    \label{fig3}
\end{figure}

The overall architecture of the proposed network is shown in Figure~\ref{fig3}. 
It consists of three main parts, which are the feature extractor, multi-stages attention module, and NSNP module. 
This study employs pre-trained ShuffleNet as the feature extractor. 
ShuffleNet involves three main stages of feature extraction. 
An image first is send to the network followed by initial convolution and downsampling processes. 
Then the images are sent to stage 1, 2, and 3 in sequence. 
The resolution size of the feature maps under the three stages is different, which represents the high-level semantic information at different scales. 
We arrange the spatial attention module after the output of each stage of ShuffleNet, finally fusing the attention information in the three different scales of stages.  
Furthermore, a channel attention block is set to analyse and weight the different channels. 
For further encoding and progressive downscaling of the high-dimensional semantic information to the low-dimensional categorical information, the feature map with the fused spatial and channel attention information is fed into the NSNP module. 
For different tasks, the fully connected layers can be directly addressed corresponding to the number of classifications.

\subsection{NSNP module embedded in pre-trained DCNN}

\begin{figure}
    \centering
    \includegraphics[width=4in]{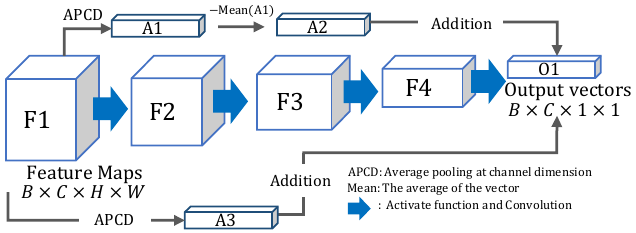}
    \caption{NSNP module. F1-F4 are feature maps with four dimensions. All A1, A2, A3, and O1 are vectors with $C$ elements.}
    \label{fig4}
\end{figure}

The NSNP module incorporates both encoding and dimensionality reduction processes in the network. 
As shown in Figure~\ref{fig3}, the NSNP module is placed after the feature extraction, where $U$ and $g(U)$ denote the state value of the last temporal state of the NSNP neuron with autapse and the state value consumed by the current activation, respectively. Its detailed construction is shown in Figure~\ref{fig4}. We adopt three consecutive layers of convolution to downsample the output image from the initial feature maps $F1$ of size $C*H*W$ to $C*1*1$. In the convolutional layers, we follow Eq.~(\ref{eq3}) with the input feature maps first with nonlinear activation and then with convolution manipulation. As shown in Figure~\ref{fig4}, Given input feature maps $F1 \in R^{C \times H \times W}$, three convolutions yield output vector $O1$. $A2$ and $A3$ are also two vectors. $A3$ is obtained by globally averaging the channel dimensions of $F1$. $A1$ and $A3$ are equal in the neural network. We take the average of the elements in $A3$. With subtracting its mean value, $A1$ retains his deviation, i.e., $A2$. Finally, $A2$ and $A3$ are added to $O1$. The mechanism of NSNP module can be summarized in the following equation:
\begin{equation}
    O1=W_3 * \delta(W_2 * \delta(W_1 * \delta(F1)) + 2 \times \varphi^{(s)}_{avg}(F1) - \varphi^{(c)}_{avg}(A1)
\end{equation}
where $\varphi^{(c)}_{avg}$ and $\varphi^{(s)}_{avg}$ are global average pooling at channel dimension and spatial dimension, respectively. $\delta$ is ReLU function. $W_1, W_2$, and $W_3$ are all metrixes, which stands the convolution process with different strides. Therefore, it is sufficient to set up a fully connected layer of $C$ channels to 8 or 2 for mapping $C$ classifications to different ones.

\begin{figure}
    \centering
    \includegraphics[width=4in]{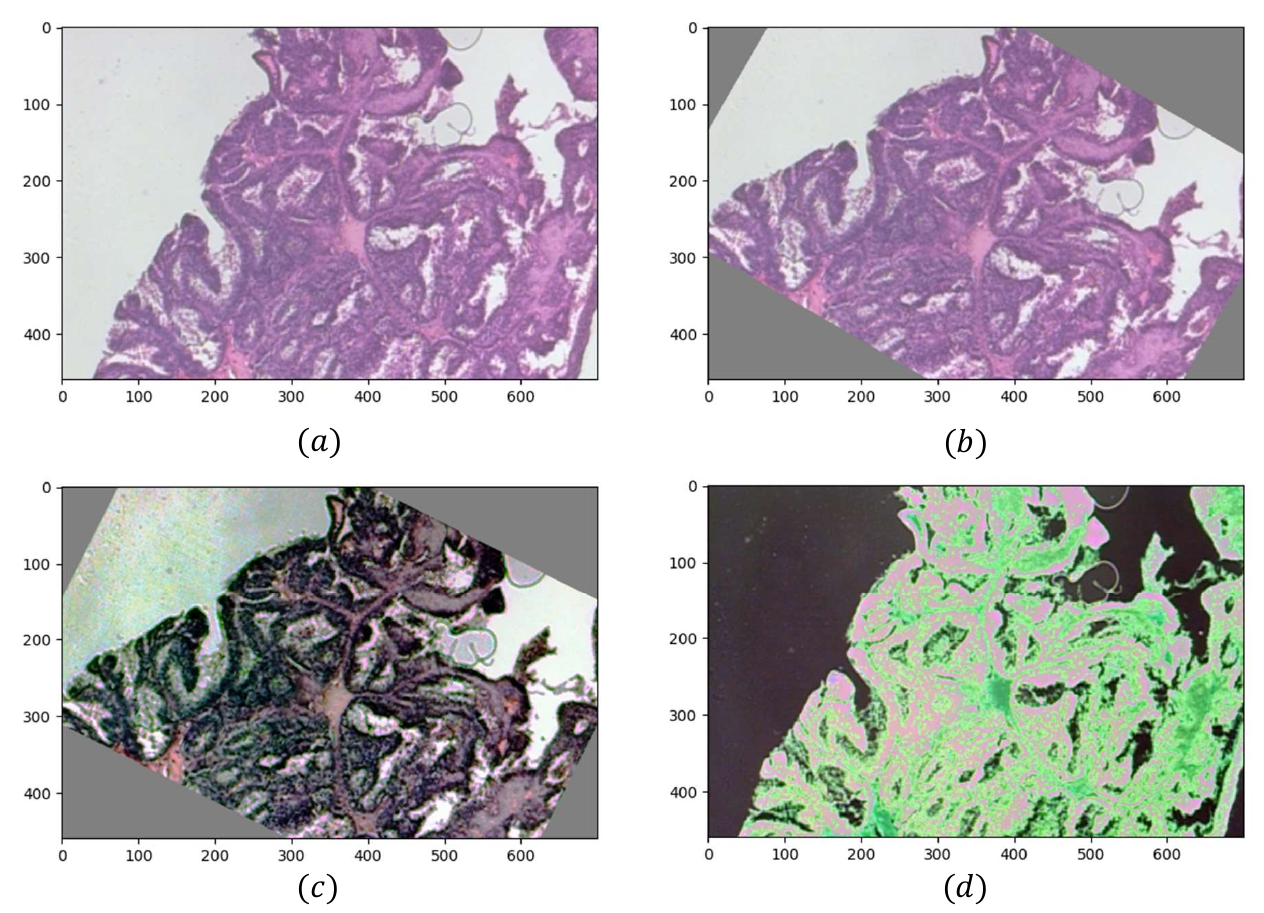}
    \caption{Data enhancement techniques of different types. (a) Original image;(b) Rotation; (c) Histogram equalization and rotation; (d) Solarization and automatic contrast enhancement.}
    \label{fig5}
\end{figure}

\subsection{Pre-processing}

Though BreakHis provides a substantial number of images for breast cancer analysis, it may not be sufficient for large-scale machine-learning models.
Therefore, we employ data augmentation techniques to increase training dataset. 
Rotating and flipping is an effective way to quickly add large amounts of data. To enhance the image texture features, we also perform histogram equalization, gamma correction, etc. Some of the enhanced images are shown in Figure~\ref{fig5}. It is worth mentioning that both the enhanced data and the original image are fed into the model for training, which improves the generalization ability of the model. These methods help to prevent overfitting. 
Expanding dataset also helps to balance the distribution of different classes. 
Before being fed into the model for training, the pixel values of the image are scaled to values between 0 and 1.

\begin{figure}
    \centering
    \includegraphics[width=4in]{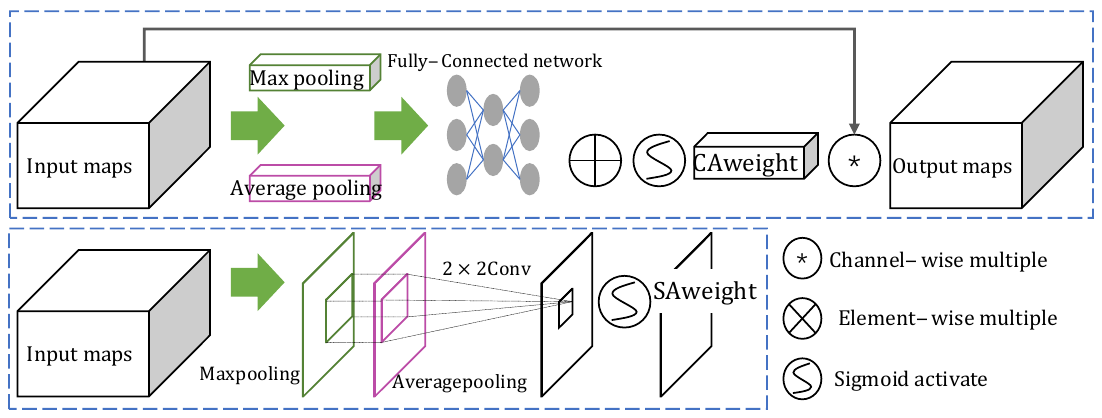}
    \caption{Two attention modules: the above is channel attention (CA) and the below is spatial attention (SA).}
    \label{fig6}
\end{figure}

\subsection{multi-stagesattention framework}

The convolutional block attention module (CBAM)~\cite{woo2018cbam} is an attention mechanism that focus attention on salient feature to improve the performance of CNNs. 
The CBAM block consists of two sub-blocks: the channel attention(CA) block and the spatial attention(SA) block. 
As shown in Figure~\ref{fig6}, the CA block assigns a weight value to each channel by computing global information for each channel. 
This method allows a model to learn the trait of different channels adaptively. 
Similarly, the SA block determines the weight value for each spatial location by learning the spatial relationship of each feature map. 

On CA stages, the feature maps $F \in R^{C \times H \times W}$ are turned into two vectors by max pooling $\varphi^{(c)}_{max}$ and average pooling 
$\varphi^{(c)}_{avg}$. 
Through a fully connected layer with shared weights $w_{fc}$, the information of two vectors is rearranged. 
The two sets of channel information are directly summed up, followed by nonlinear activation $\delta$. 
The CAweight received is a set of vectors with focused information, which can be assigned to the original feature maps channel attention by multiplying it with the original feature maps. The mechanism of CA can be summarized in the following equation:
\begin{equation}
    A^{(c)}(F)=\delta (w_{fc} \times \varphi^{(c)}_{avg}(F)+w_{fc} \times \varphi^{(c)}_{max}(F))\times F
\end{equation}
where $\delta$ represents activate function.
\par SA follows a similar approach, also using max pooling $\varphi^{(s)}_{avg}$ and average pooling $\varphi^{(s)}_{avg}$ on the feature maps $F$, which differs from CA in that this pooling is conducted in the spatial dimension. 
Further, the feature maps are reduced in dimension by the convolution layer and activate function $\delta$. The mechanism of SA can be summarized in the following equation:
\begin{equation}
    A^{(s)}(F)=\delta (w * ( \varphi^{(s)}_{avg}(F) \textcircled {c} \varphi^{(s)}_{max}(F)))
\end{equation}
where $\textcircled {c}$ means channel dimensional concat, $*$ means the convolution process, and $w$ is a metrix with 4 elements.
\par ShuffleNet consists of three main stages, where different attention information matrices can be available at different stages. This facilitates us to fuse the attention information of different stages. 
To obtain attention information at different scales, we apply SA to each stage to obtain attention information SAweight. 
The size of the SAweight for different stages is unequal, whose ratio of the output map size for different stages is 4:2:1. 
Therefore, single map downsampling is used to adjust the size of the feature maps adaptively. 
With three stages, three spatial attention information maps are available. 
The single map downsampling and multi-attention adaptive fusion are generated using $2 \times 2$ convolution and $3 \times 3$ convolution, respectively. Let the outputs of stage 1, 2, and 3 be $F_1$, $F_2$, and $F_3$, respectively. The attention maps $A^{(s)}_1$, $A^{(s)}_2$, and $A^{(s)}_3$ at different scales are obtained after passing through the respective SA modules. The process of fusion at different scales can be expressed by the following equation:
\begin{equation}
    A^{(s)}_{fusion}=\left ((A^{(s)}_1*w_1) \textcircled{c} (A^{(s)}_2*w_2) \textcircled{c} A^{(s)}_3 \right)*w_{fusion}
\end{equation}
where $A^{(s)}_{fusion}$ is the final attention map. $w_{fusion}, w_1$ and $w_2$ are all matrixes, where $w_{fusion}$ has 9 elements and the other two have both 4 elements. 
To enable NSNP to selectively focus on certain channel information, CA is applied before the NSNP module.

\section{Experimental results}\label{sec4}

\subsection{Dataset}
\begin{figure}[!h]
    \centering
    \includegraphics[width=4.5in]{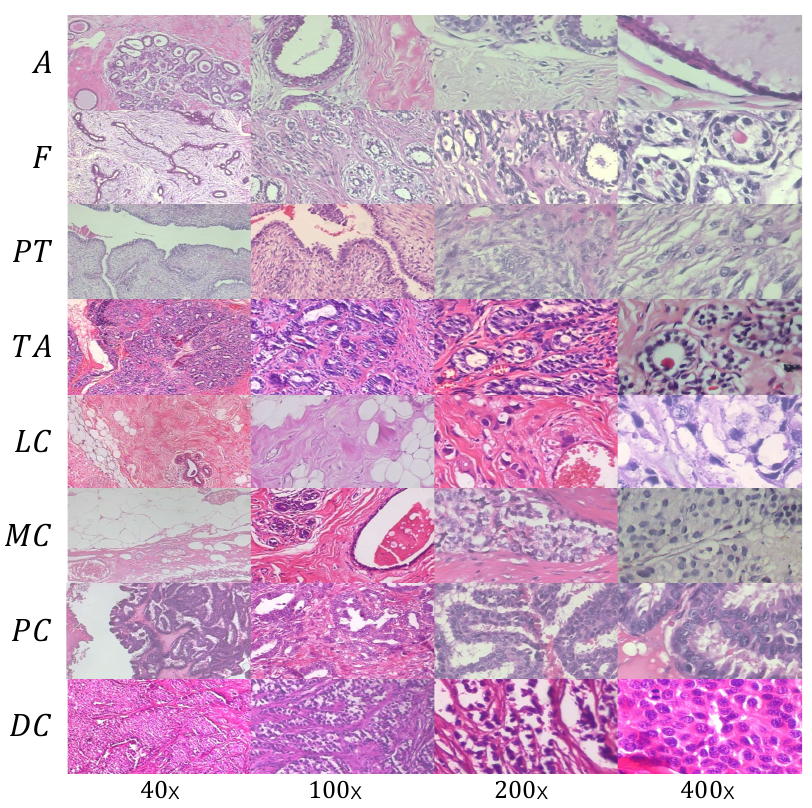}
    \caption{histological images from BreakHis dataset. Different kinds of histopathology images at different magnifications are randomly selected from the dataset.}
    \label{fig7}
\end{figure}

Breast cancer is one of the most common types of cancer affecting women worldwide. Early detection is crucial for treatment. 
Among the various imaging techniques available for breast cancer detection, histological pathology images are widely considered the golden standard for medical image analysis~\cite{debelee2020survey}.  
In this study, we aim to address the use of histological images for breast cancer diagnosis and treatment. 
To this end, we utilize the publicly available BreakHis dataset, which contains 7909 images of breast tissue. 
The dataset is divided into two main categories, benign and malignant tumors, each of which contains four subcategories. 
The benign tumors include adenosis (A), fibroadenoma (F), phyllodes tumor (PT), and tubular adenoma (TA). 
The malignant tumors include ductal carcinoma (DC), lobular carcinoma (LC), mucinous carcinoma (MC), and papillary carcinoma (PC). 
These subcategories are based on the specific cellular structures and characteristics of the tumors. 
The tumor images in the BreakHis dataset are captured at different magnifications, 40x, 100x, 200x, and 400x. 
The details of some histopathograms are shown in Figure~\ref{fig7}.

\subsection{Evaluation criterions}

In this paper, three metrics are applied to the classification task on the dataset, including accuracy, recall (same as sensitivity), and precision. 
Among them, true positive (TP), true negative (TN), false positive (FP), and false negative (FN) are used to denote predicted correct benign, predicted correct malignant, predicted incorrect benign, and predicted incorrect malignant, respectively. 
The three metrics are given by 
\begin{equation}
    Accuracy = \frac{TP+TN}{TP+TN+FP+FN}
\end{equation}
\begin{equation}
    Precision = \frac{TP}{TP+FP}
\end{equation}
\begin{equation}
    Recall = \frac{TP}{TP+FN}
\end{equation}

\subsection{Implementation}

The proposed method is implemented on PyTorch framework. 
To ensure the convergence towards the optimal solution, we resort to the stochastic gradient descent method, whereby the cosine annealing strategy is invoked for optimization purpose~\cite{loshchilov2016sgdr}. 
In our experiments, the learning rate is changed based on a simple warm restart approach, which can be described by 
\begin{equation}
    \zeta_t = \zeta^i_{min}+\frac{1}{2} (\zeta^i_{max}-\zeta^i_{min} ) \Big(1+cos(\frac{T_{cur}}{T_i}\pi) \Big)
\end{equation}
where $i$ indicates the $i$th execution of hot restart, $\zeta^i_{max}$ and $\zeta^i_{min}$ denote the maximum and minimum values of the learning rate, respectively; 
$T_{cur}$ and $T_{i}$ denote the number of epochs elapsed since the last hot restart and the number of epochs to be trained for the $i$th hot restart, respectively. 
To reduce the computational overhead, we set the maximum and minimum values of learning rats for each round of warm restart to the same value. 
$\zeta^i_{min}$ and $\zeta^i_{max}$ are set to 0.0005 and 0.1, respectively. 
Besides, we set the batch size to 30 and set the epoch to 600 throughout our experimentation process.

\subsection{Comparison with state-of-the-art methods}

Apart form ShuffleNet, we also conduct experiments with different backbone networks for all magnification images, including VGG16~\cite{simonyan2014very}, ResNet50~\cite{he2016deep}, and Inception~\cite{ioffe2015batch}. The results are shown in Table~\ref{tab1}. ShuffleNet achieves the best Acc, Sen, and Pre among these four backbone networks for both binary and 8 classifications. On the contrary, larger networks exhibit worse performance. Inception only achieves 90.11\% and 83.46\% Acc on the 8 classification and binary classification tasks, respectively. Therefore, ShuffleNet is chosen as the primary model in the proposed network. BreakHis dataset has two classification tasks, which are binary and eight classifications. 
The specific groupings are shown in Table~\ref{tab2}. 
In most of the literature, the network is trained separately based on single different magnifications. 
While for few papers, it is trained together with all magnifications. In this paper, we experiment with different magnifications and different classification methods for the task, with four sets of experiments in total. As a result, there are four sets of experiments, including a single magnification for the binary classification task, a multi-classification task at all magnifications, a single magnification for the multi-classification task, and a multi-classification task at all magnifications. 
The binary and multi-classification results at a single magnification are shown in Table~\ref{tab3}. 
The proposed method is compared with the state-of-the-art methods in the last 5 years, including Pratiher et al.~\cite{pratiher2018manifold}, Erfankhah et al.~\cite{erfankhah2019heterogeneity}, Sudharshan et al.~\cite{sudharshan2019multiple}, Shallu et al.~\cite{shallu2019automatic}, Li et al.~\cite{li2020classification}, Boumaraf et al.~\cite{boumaraf2021new}, and Zaalouk et al.~\cite{zaalouk2022deep}. 
Among them, both Boumaraf et al. and Zaalouk et al. provide comparative scores in both 8 classification and binary classification. 
The 8-classified Acc of the proposed method at four magnifications of 40x, 100x, 200x, and 400x are 93.69\%, 93.40\%, 94.71\%, and 91.58\%, respectively. 
The binary classification accuracies at these four magnifications are 100\%, 99.53\%, 98.56\%, and 98.42\%, respectively. 
The accuracy of the proposed method is 100\% and 99.53\% at 40x and 100x magnification in binary classification, which is the first place. 
The accuracy achieves 94.71\% and 91.58\% at 200x and 400x magnification in multi-classification respectively, which is also the first place. 
Zaalouk et al. achieve 100\% and 99.46\% accuracy in binary classification at 200x and 400x, which surpasses ours by 1.44\% and 1.04\%, respectively. 
However, the multi-classification of their method is lower than ours for the two magnifications of 200x and 400x.
The binary classification and multi-classification results of images at all magnifications are shown in Table~\ref{tab4}. 
The proposed method is compared with four methods, including Kassani et al.~\cite{kassani2019breast}, Celik et al.~\cite{celik2020automated}, Boumaraf et al.~\cite{boumaraf2021new}, and Karthik et al.~\cite{karthik2022classification}. 
For Boumaraf et al.'s method, we have compared the experiments of single magnification images in Table~\ref{tab3}. 
As shown in Table~\ref{tab4}, the proposed method achieves first place in the 8 classified ones, with 96.32\%, 96.54\%, and 96.23\% for Acc, Sen, and Pre, respectively. 
On the binary classification task, the proposed method just achieves first place on Pre, surpassing the second place kar by 0.06\%. 
With Acc and Sen, it falls behind Kar and Boun with 0.41\% and 0.05\%, respectively.

\begin{table}[ht]

\caption{Results of employing NSNP neurons with autapses on other popular models}
\label{tab1}
\begin{tabular}{ccccc}
\hline
Backbone                     & \begin{tabular}[c]{@{}c@{}}Classification\\ Task\end{tabular} & Acc     & Sen     & Pre     \\ \hline
\multirow{2}{*}{Inception \cite{ioffe2015batch}} & 8                                                             & 90.11\% & 88.76\% & 88.84\% \\
                             & 2                                                             & 83.46\% & 74.25\% & 88.09\% \\
\multirow{2}{*}{ResNet50 \cite{he2016deep}}    & 8                                                             & 92.71\% & 91.03\% & 91.95\% \\
                             & 2                                                             & 91.63\% & 89.88\% & 90.52\% \\
\multirow{2}{*}{VGG16 \cite{simonyan2014very}}       & 8                                                             & 93.73\% & 92.18\% & 93.71\% \\
                             & 2                                                             & 97.21\% & 97.03\% & 96.53\% \\ 
\multirow{2}{*}{ShuffleNet \cite{zhang2018shufflenet}}       & 8                                                             & 96.32\% & 96.54\% & 96.23\% \\
                             & 2                                                             & 99.14\% & 98.96\% & 99.06\% \\ \hline
\end{tabular}
\end{table}

\begin{table}[h]
	\caption{Distribution of images in BreakHis dataset}
	\label{tab2}
	\begin{tabular}{lllllllllll}
		\hline
		\multicolumn{2}{l}{\multirow{2}{*}{}} & \multicolumn{4}{l}{Malignant} & \multicolumn{4}{l}{Benign} & Sum  \\ \cline{3-10}
		\multicolumn{2}{l}{}                  & PC    & MC    & LC    & DC    & TA   & PT   & F     & A    &      \\ \cline{3-10}
		\multirow{4}{*}{BreakHis}    & 40x    & 145   & 205   & 156   & 864   & 149  & 109  & 253   & 114  & 1995 \\
		& 100x   & 142   & 222   & 170   & 903   & 150  & 121  & 260   & 113  & 2081 \\
		& 200x   & 135   & 196   & 163   & 896   & 140  & 108  & 264   & 111  & 2013 \\
		& 400x   & 138   & 169   & 137   & 788   & 130  & 115  & 237   & 106  & 1820 \\
		Sum                          &        & 560   & 792   & 626   & 3451  & 569  & 453  & 1014  & 444  & 7909 \\ \hline
	\end{tabular}
\end{table}

\begin{table}[ht]
\caption{Comparison of experiments at single magnifications with state-of-the-art methods}
\label{tab3}
\resizebox{\linewidth}{!}{
\begin{tabular}{ccccccc}
\hline
\multirow{2}{*}{Method}          & \multirow{2}{*}{\begin{tabular}[c]{@{}c@{}}Published\\ Year\end{tabular}} & \multirow{2}{*}{\begin{tabular}[c]{@{}c@{}}Classification\\ Task\end{tabular}} & \multicolumn{4}{c}{Acc}                                                   \\ \cline{4-7} 
                                 &                                                                           &                                                                                & 40x              & 100x             & 200x             & 400x             \\ \hline
Pratiher et al. \cite{pratiher2018manifold}                        & 2018                                                                      & 2                                                                              & 96.8\%           & 98.1\%           & 98.2\%           & 97.5\%           \\
Erfankhah et al. \cite{erfankhah2019heterogeneity}                 & 2018                                                                      & 2                                                                              & 88.3\%           & 88.3\%           & 87.1\%           & 83.4\%           \\
Sudharshan et al. \cite{sudharshan2019multiple}                & 2019                                                                      & 2                                                                              & 87.8\%           & 85.6\%           & 80.8\%           & 82.9\%           \\
Shallu et al. \cite{shallu2019automatic}                   & 2019                                                                      & 2                                                                              & 90.4\%           & 86.3\%           & 83.1\%           & 81.3\%           \\
Li et al. \cite{li2020classification}                      & 2020                                                                      & 2                                                                              & 89.1\%           & 85.0\%           & 87.0\%           & 84.5\%           \\
\multirow{2}{*}{Boumaraf et al. \cite{boumaraf2021new}} & \multirow{2}{*}{2021}                                                     & 8                                                                              & 94.49\%          & 93.27\%          & 91.29\%          & 89.56\%          \\
                                 &                                                                           & 2                                                                              & 99.25\%          & 99.04\%          & 99\%             & 98.08\%          \\
\multirow{2}{*}{Zaalouk et al. \cite{zaalouk2022deep}}  & \multirow{2}{*}{2022}                                                     & 8                                                                              & \textbf{97.01\%} & \textbf{95.17\%} & 91.54\%          & 90.22\%          \\
                                 &                                                                           & 2                                                                              & \textbf{100\%}   & 99.52\%          & \textbf{100\%}   & \textbf{99.46\%} \\
\multirow{2}{*}{The proposed}    & \multirow{2}{*}{}                                                         & 8                                                                              & 93.69\%          & 93.40\%          & \textbf{94.71\%} & \textbf{91.58\%} \\
                                 &                                                                           & 2                                                                              & \textbf{100\%}   & \textbf{99.53\%} & 98.56\%          & 98.42\%          \\ \hline
\end{tabular}}
\end{table}

\begin{table}[]
\caption{comparison of experiments at all magnifications with state-of-the-art methods}
\label{tab4}
\resizebox{\linewidth}{!}{
\begin{tabular}{cccccc}
\hline
Method                           & \begin{tabular}[c]{@{}c@{}}Published\\ Year\end{tabular} & \begin{tabular}[c]{@{}c@{}}classification\\ Task\end{tabular} & Acc              & Sen              & Pre              \\ \hline
Kassani et al. \cite{kassani2019breast}                  & 2019                                                     & 2                                                             & 98.13\%          & 98.54\%          & 98.75\%          \\

Celik et al. \cite{celik2020automated}                    & 2020                                                     & 2                                                             & 91.96\%          & 93.64\%          & 94.58\%          \\
\multirow{2}{*}{Boumaraf et al. \cite{boumaraf2021new}} & \multirow{2}{*}{2021}                                    & 8                                                             & 92.03\%          & 90.28\%          & 91.39\%          \\
                                 &                                                          & 2                                                             & 98.42\%          & \textbf{99.01\%} & 98.75\%          \\
                                 Karthik et al. \cite{karthik2022classification}                 & 2022                                                     & 2                                                             & \textbf{99.55\%} & 99\%             & 99\%             \\
\multirow{2}{*}{The proposed}    & \multirow{2}{*}{}                                        & 8                                                             & \textbf{96.32\%} & \textbf{96.54\%} & \textbf{96.23\%} \\
                                 &                                                          & 2                                                             & 99.14\%          & 98.96\%          & \textbf{99.06\%} \\ \hline
\end{tabular}}
\end{table}

\begin{figure}[]
    \centering
    \includegraphics[width=4.5in]{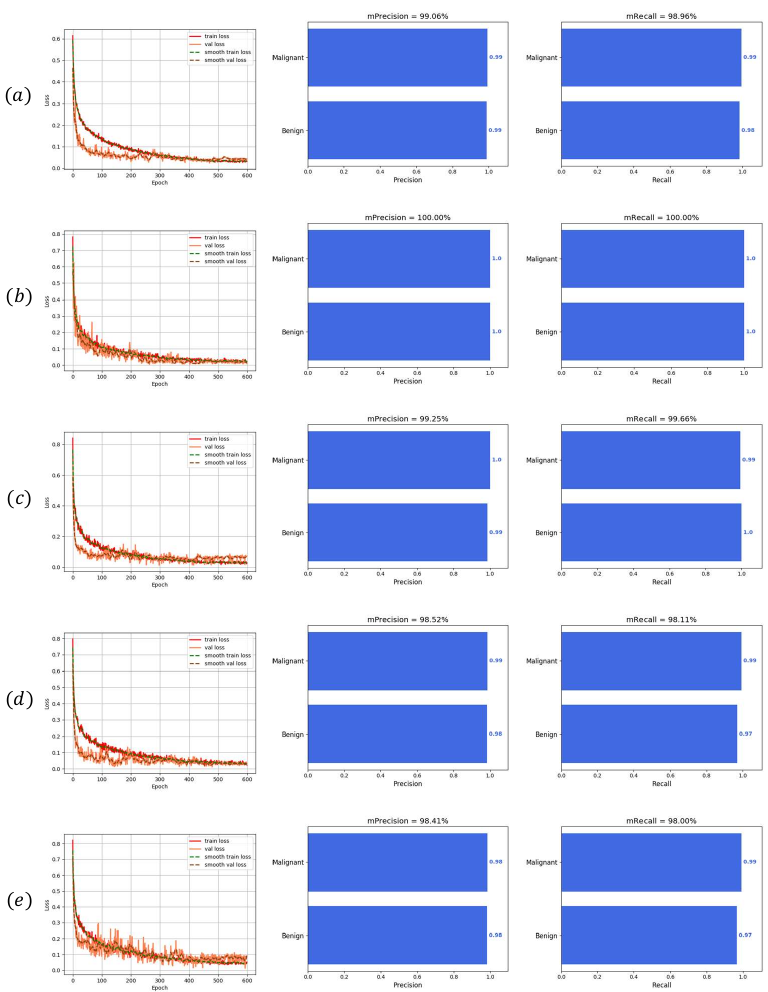}
    \caption{All the binary classification training results are plotted. The first column shows the result of decreasing loss with the number of epochs, and the two bars in the second and third columns are the average Precision and Recall(same as Sensitivity) of Benign and malignant, respectively. (a)All magnification,(b)40x,(c)100x,(d)200x,(e)400x.}
    \label{fig8}
\end{figure}

\begin{figure}[]
    \centering
    \includegraphics[width=4.5in]{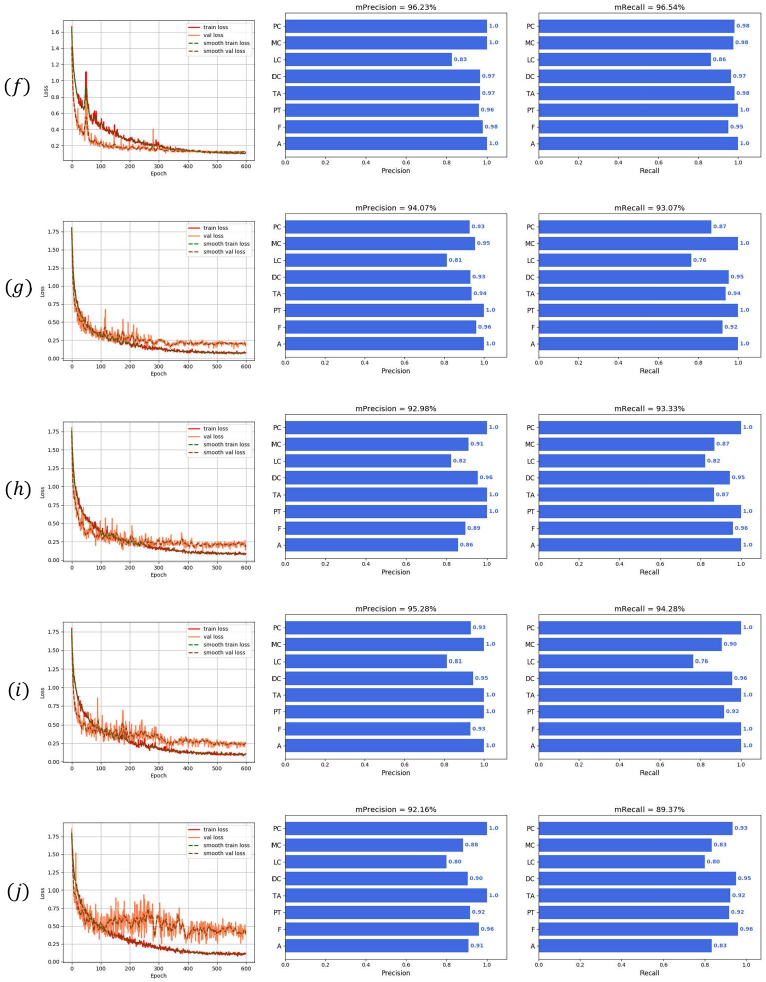}
    \caption{All the multi-classification training results are plotted. The first column shows the result of decreasing loss with the number of epochs, and the two bars in the second and third columns are the average Precision and Recall(same as Sensitivity) of adenosis, fibroadenoma, phyllodes tumor, and tubular adenoma, conducted carcinoma, lobular carcinoma, mucinous carcinoma, and papillary carcinoma, respectively. (a)All magnification,(b)40x,(c)100x,(d)200x,(e)400x.}
    \label{fig9}
\end{figure}

\subsection{Visualization analysis}

The training procedures of the experiments are an essential aspect of this study and are depicted clearly in Figures~\ref{fig8} and \ref{fig9}. 
The first column of Figures~\ref{fig8} and ~\ref{fig9} both provide information on the loss with epoch. Here, the red curve corresponds to the training loss, while the orange curve represents the validation loss; 
the second and third columns of the figure exhibit the precision and recall on the test set respectively, with (a)-(e) representing all magnifications, including 40x, 100x, 200x, and 400x respectively. 
Figure~\ref{fig8} illustrates the training procedures for the binary classification tasks. 
It is evident from the loss results that the overall training process is quite smooth. 
The loss trend is observed to decrease gradually, without many ups and downs. 
As can be seen from Figure~\ref{fig8}(a), both precision and recall achieve an accuracy of more than 98\% on two different classes. 
Even though the entire dataset is divided into 5429 malignant and 2480 benign classifications, with a large discrepancy between both, the final classification accuracy of both groups is almost unaffected. 
Figure~\ref{fig8}(b) shows that the average precision and recall both achieve 100\%. 
Each classification in (c) achieves 99\% or more for both precision and recall. 
It achieves 98\% or more in (d) and (e) on average. 
For the same dataset, the model converges harder as the classification types increase. 
Compared with the convergence effect of loss in Figure~\ref{fig8}, the results shown in Figure~\ref{fig9} appear to be more variable, especially the validation loss in column (j), which varies between 100 and 400 and is difficult to converge. 
Figure~\ref{fig9} highlights that LC has a crucial impact on the average accuracy of both precision and recall. 
This observation is noteworthy given that the multi-classification dataset features a distribution of various types of pathology images that are not evenly dispersed. 
From the various types of data in Table~\ref{tab2}, we rank the data volume from least to most as A, PT, PC, TA, LC, MC, F, and DC. 
On the other hand, the amount of data LC is not the least. 
However, the overall data distribution shows that the recognition rate of LC is certainly the worst. 
Above this, it is observed that the identification of pathology type and data amount is not exactly positively correlated.

\begin{table}[h]

	\caption{Ablation at all magnifications}
	\label{tab5}
	\begin{tabular}{llll}
		\hline
		Method            & Acc     & Sen     & Pre     \\ \hline
		Backbone          & 95.47\% & 95.16\% & 95.65\% \\
		Backbone+NSNP     & 95.83\% & 95.65\% & 95.52\% \\
		Backbone+MSA      & 95.96\% & 96.04\% & 96.04\% \\
		Backbone+NSNP+MSA & 96.32\% & 96.54\% & 96.23\% \\ \hline
	\end{tabular}
\end{table}

\subsection{Ablation experiments}

In ablation experiments, we adopt ShuffleNet as the backbone network. Two separate experiments are conducted using NSNP and MSA modules individually. All magnifications of the images are used to ensure accurate and reliable results during testing. 
As present in Table~\ref{tab5}, the addition of NSNP and MSA modules results in significant improvements to the accuracy of the entire network. 
The direct employment of ShuffleNet network classification yields 95.47\%, 95.16\%, and 95.65\% improvement effects on Acc, Sen, and Pre, respectively. 
After applying NSNP and MSA, Acc increases to 95.83\% and 95.96\% respectively. 
Specifically, the Acc of the whole network is strengthened by 0.36\% and 0.49\%, respectively. 
Although the effect may appear to be minimal, it is noteworthy that when both modules were attached to the network, the Acc rises by 0.95\%. 
On the other hand, Sen and Pre increases by 1.38\% and 0.58\%, respectively. 
NSNP module works as a role who makes a dimension reduction and classification in the network. Attention mechanism can prioritize and filter out the more important information to make the NSNP work more precisely. 
The result indicates that both NSNP and MSA modules are effective in enhancing the performance of network.

\section{Conclusions}\label{sec5}

In this paper, we introduce a new type of neuron, called NSNP neuron with autapses and give its mathematical model.
According to the mathematical definition, we construct NSNP module to further encode the feature information extracted by CNNs. 
In addition, we propose a multi-stagesattention framework that fuses feature information at different scales. To transfer image styles from ImageNet to histopathological images, we employ ShuffleNet as the basic feature extractor. We also introduce cosine annealing to adaptively adjust the learning rate, which prevents network from falling into local optima. The evaluations are conducted on the BreakHis dataset, which is a comprehensive collection of breast histopathological images that provide not only binary classification tasks but also multi-classification tasks. We explore binary and multiclassification tasks for images at both different and same magnification. The experimental results confirm the effectiveness of the proposed method. The model shows the competitive performance among the 11 most advanced methods compared. The binary classification accuracy for all magnifications resides around 99\%. Ablation experiments demonstrates the effectiveness of the proposed NSNP neurons with autapses combined with MSA. In future work, we will pay more attention on studying methods through lager neural networks. 

%\begin{acknowledgements}
%This work was partially supported by National Natural Science Foundation of China (No. 62076206 and No. 62176216), China.
%\end{acknowledgements}

% Authors must disclose all relationships or interests that 
% could have direct or potential influence or impart bias on 
% the work: 
%
 \section*{Conflict of interest}

Authors have no conficts of interest to declare.

 \section*{Funding}

\section*{Ethical Approval}

Yes.

\section*{Consent to Participate}

None.

\section*{Author Contributions}

All authors contributed to the study conception and design. All authors read and approved the final manuscript.

\section*{Data Availability Statement}

The data that support the findings of this study are openly available in reference \cite{spanhol2015dataset}.

%% If you have bibdatabase file and want bibtex to generate the
%% bibitems, please use
%%
 \bibliographystyle{elsarticle-num}

%% else use the following coding to input the bibitems directly in the
%% TeX file.

% \begin{thebibliography}{00}

% %% \bibitem{label}
% %% Text of bibliographic item

% \bibitem{}

% \end{thebibliography}
\end{document}